\documentclass[letterpaper]{article} 
\usepackage{aaai2026}  
\usepackage{times}  
\usepackage{helvet}  
\usepackage{courier}  
\usepackage[hyphens]{url}  
\usepackage{graphicx} 
\usepackage{natbib}  
\usepackage{caption} 
\usepackage{algorithm}
\usepackage{algorithmic}
\usepackage{amsmath}
\usepackage{amsfonts}
\usepackage{booktabs}
\usepackage{enumitem}
\usepackage{float}
\usepackage{newfloat}
\usepackage{listings}
\usepackage{pifont}  

\DeclareCaptionStyle{ruled}{labelfont=normalfont,labelsep=colon,strut=off} 
\floatstyle{ruled}
\newfloat{listing}{tb}{lst}{}
\floatname{listing}{Listing}
\title{MD-RWKV-UNet: Scale-Aware Anatomical Encoding with Cross-Stage Fusion for Multi-Organ Segmentation}
\author {
    Zhuoyi Fang
}
\affiliations{
    School of Mathematical Sciences\\
    Yangzhou University\\
    230802107@stu.yzu.edu.cn
}

\usepackage{bibentry}

\begin{document}

\maketitle

\begin{abstract}
Multi-organ segmentation in medical imaging remains challenging due to large anatomical variability, complex inter-organ dependencies, and diverse organ scales and shapes. Conventional encoder-decoder architectures often struggle to capture both fine-grained local details and long-range context, which are crucial for accurate delineation—especially for small or deformable organs. To address these limitations, we propose MD-RWKV-UNet, a dynamic encoder network that enables scale-aware representation and spatially adaptive context modeling. At its core is the MD-RWKV block, a dual-path module that integrates deformable spatial shifts with the Receptance Weighted Key Value mechanism, allowing the receptive field to adapt dynamically to local structural cues. We further incorporate Selective Kernel Attention to enable adaptive selection of convolutional kernels with varying receptive fields, enhancing multi-scale interaction and improving robustness to organ size and shape variation. In parallel, a cross-stage dual-attention fusion strategy aggregates multi-level features across the encoder, preserving low-level structure while enhancing semantic consistency. Unlike methods that stack static convolutions or rely heavily on global attention, our approach provides a lightweight yet expressive solution for dynamic organ modeling. Experiments on Synapse and ACDC show state-of-the-art performance, particularly in boundary precision and small-organ segmentation.

\end{abstract}

\section{Introduction}

Medical image segmentation plays a critical role in modern clinical workflows, serving as a fundamental step in many image-guided diagnostic and treatment planning tasks \cite{Patil2013,Ramesh2021,Pham2000}. With the rise of data-driven approaches, deep learning has become the dominant paradigm in automated medical image segmentation, offering significant advantages in accuracy, speed, and generalization. Among various architectures, convolutional neural networks (CNNs) \cite{Liu2021,Hesamian2019,Razzak2017,Gao2025}, particularly UNet \cite{Ronneberger2015} and its derivatives, have shown remarkable success due to their ability to model local structures and leverage translational invariance. However, the intrinsic locality of CNNs limits their capacity to capture global semantic context, which is essential for accurate delineation in complex anatomical scenes \cite{Wang2022,Norouzi2014}.

To address these limitations, recent efforts have explored sequence modeling frameworks based on linear attention mechanisms, such as Mamba \cite{gu2023} and RWKV \cite{peng2023}. These models provide efficient and scalable alternatives to traditional self-attention by enabling long-range dependency modeling with linear computational complexity. Their ability to maintain contextual coherence across spatially distant regions makes them well-suited for medical image analysis tasks, particularly in high-resolution or multi-organ settings.

Despite these advances, multi-organ segmentation remains especially challenging due to large anatomical variability, complex inter-organ spatial relationships, and significant diversity in organ size, shape, and texture. Accurate segmentation in such settings requires models that can not only retain fine-grained boundary details but also adaptively capture long-range dependencies across scales and structures.

To tackle these challenges, we propose MD-RWKV-UNet, a novel dynamic encoder framework tailored for multi-organ segmentation. Our architecture integrates three key innovations: (1) the MD-RWKV block, which combines deformable spatial shifts with the Receptance Weighted Key Value (RWKV) mechanism to support efficient long-range context modeling with spatial adaptability; (2) Selective Kernel Attention (SKAttention), which dynamically adjusts receptive fields to better accommodate organ scale and shape variation; and (3) a cross-stage dual-attention fusion strategy that adaptively aggregates hierarchical features, preserving structural detail while enhancing semantic consistency. These components work in concert to provide a lightweight yet expressive solution, delivering robust performance across diverse segmentation scenarios involving complex organ boundaries and scale heterogeneity.

Our main contributions are summarized as follows:
\begin{enumerate}[label=\arabic*)]
\item We propose MD-RWKV-UNet, a dynamic encoder framework for multi-organ segmentation that adaptively balances fine-grained local detail and long-range contextual information. The architecture is designed to handle anatomical variability and scale diversity by dynamically adjusting its spatial modeling behavior across encoding stages.
\item We introduce a set of novel modules to enhance scale-aware and structure-adaptive representation. Specifically, the MD-RWKV block combines deformable spatial shifts with RWKV for efficient context modeling; SKAttention enables adaptive receptive field adjustment through dynamic kernel selection; and a dual-attention cross-stage fusion mechanism integrates hierarchical features to preserve detail while enhancing semantic coherence.
\item Extensive experiments on the Synapse and ACDC datasets demonstrate that our method achieves state-of-the-art or competitive performance across a variety of organs, with notable improvements in boundary accuracy and segmentation of small or highly deformable structures.
\end{enumerate}

\section{Related work}

\subsection{Multi-organ segmentation} 
\label{subsec:segmentation}

Fully supervised multi-organ segmentation relies on accurate voxel-wise annotations. Early methods predominantly utilized convolutional neural networks (CNNs), which extract local features from medical images. For instance, Ibragimov and Xing \cite{ibragimov2017} trained separate CNNs for different head-and-neck organs, achieving good results on large structures like the spinal cord, but underperforming on smaller or irregular ones like the optic chiasm. Later, fully convolutional networks (FCNs) improved spatial consistency and efficiency by enabling end-to-end training without fully connected layers. Architectures like U-Net \cite{Ronneberger2015} and V-Net further advanced the field through encoder-decoder designs with skip connections, while nnU-Net introduced an auto-configuration pipeline that adapts to different datasets.

To address class imbalance and structural constraints, generative adversarial network (GAN)-based approaches have been introduced \cite{goodfellow2014}. For example, adversarial models combining U-Net generators with FCN discriminators improved segmentation of chest organs \cite{feng2020}, while SC-GAN incorporated shape constraints for better anatomical consistency. Despite performance gains, GANs are computationally expensive and often unstable during training, with limited success on irregular or fine-grained structures.

Recognizing CNNs’ limitations in modeling long-range dependencies, transformer-based methods such as Swin-UNet \cite{cao2022} and TransUNet \cite{chen2021} were proposed. These models effectively capture global context via self-attention mechanisms. However, transformers alone may struggle with fine boundary localization and small organ segmentation due to resolution constraints. Consequently, hybrid architectures like UNETR \cite{hatamizadeh2022itunet, zhou2021unetr} and ITUnet \cite{zhou2021} combine CNNs and transformers to exploit both local detail and global relationships, achieving more robust results across organs of various sizes.

To better handle small or overlapping structures, cascaded segmentation frameworks have emerged. These include coarse-to-fine strategies, where an initial segmentation is refined in later stages (e.g., RSTN \cite{wang2020}), and localization-to-segmentation pipelines like FocusNet \cite{li2021}, which first detect regions of interest before precise segmentation. While effective, these methods demand high computational resources and rely heavily on accurate localization in early stages.

Recent directions in multi-organ segmentation highlight the importance of generalization and efficiency. Techniques such as self-supervised learning, domain adaptation, and uncertainty modeling help reduce dependency on labeled data and improve robustness across domains \cite{zhou2022}. Meanwhile, innovations like graph neural networks (GNNs) \cite{song2021} and probabilistic priors offer new ways to encode anatomical structure. Nevertheless, challenges persist in balancing accuracy, efficiency, and interpretability—particularly in clinical deployment scenarios.

\begin{figure*}[htbp]
\centering
\includegraphics[width=\textwidth]{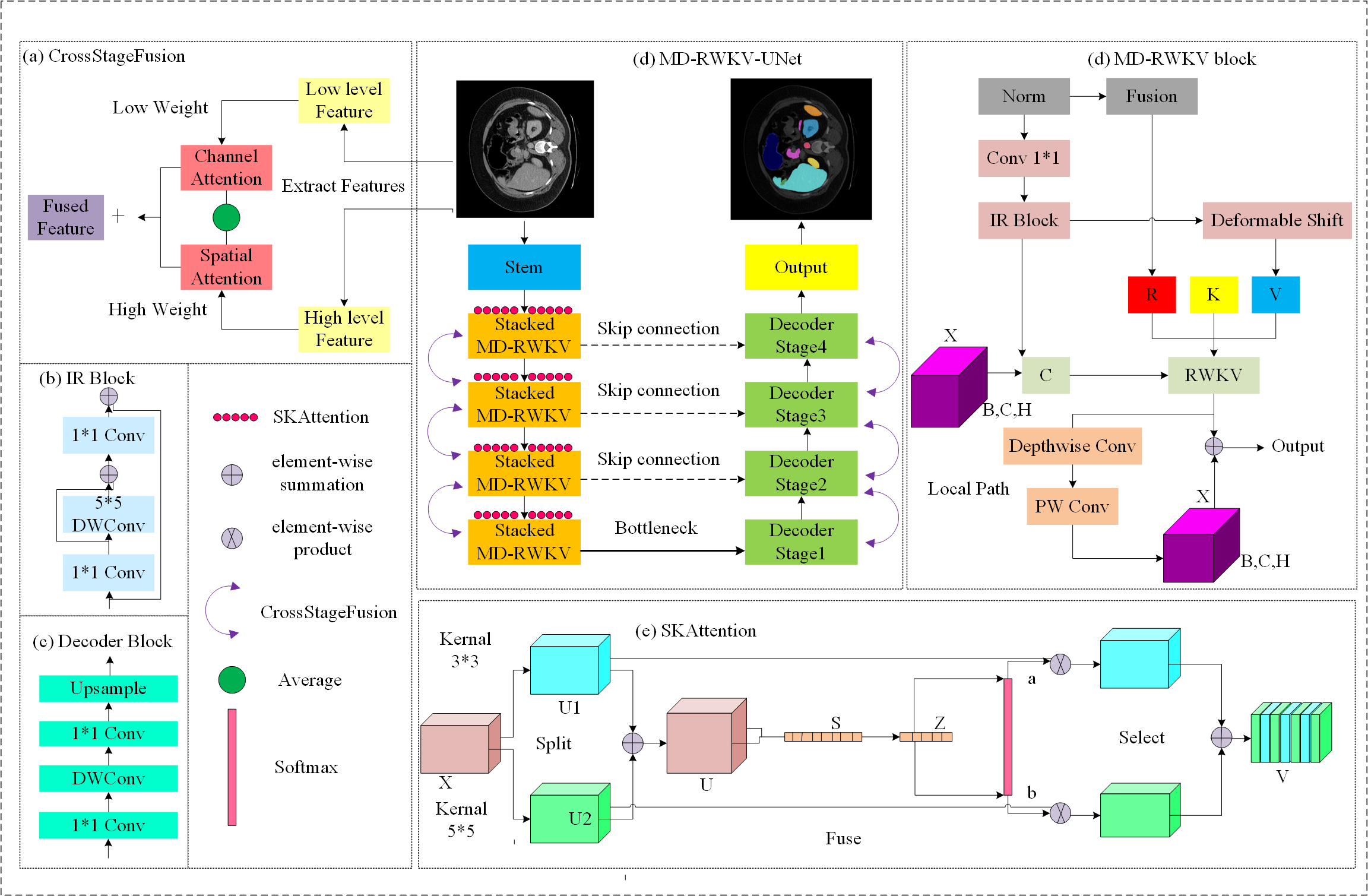}
\caption{Overall architecture of the proposed MD-RWKV-UNet.}
\label{fig:Fig1}
\end{figure*}

\subsection{RWKV and vision-RWKV}
\label{subsec:RWKV}

Recent advances in sequence modeling have sparked the development of efficient alternatives to traditional transformer architectures, with the Receptance Weighted Key Value (RWKV) model emerging as a notable breakthrough \cite{peng2023rwkv}. As a hybrid of recurrent neural networks (RNNs) and transformers, RWKV combines the linear-time efficiency of RNNs with the ability of transformers to capture long-range dependencies. This innovative design replaces self-attention with a time-mixing mechanism and gated token processing, which significantly reduces the computational and memory overhead typically associated with transformer models. As RWKV matures, researchers have begun adapting it to vision tasks through the development of Vision-RWKV (VRWKV), which integrates spatial-mix and channel-mix blocks to handle two-dimensional image data effectively. This adaptation has paved the way for incorporating RWKV into various high-resolution visual processing applications, including medical image analysis.

One of the earliest and most significant applications of RWKV in medical imaging was introduced in 2025 with the proposal of RWKV-UNet \cite{jiang2025rwkvunet}. This model integrates the RWKV block into a U-Net framework to enhance contextual understanding across spatial dimensions. In RWKV-UNet, the encoder replaces conventional convolutional layers with IR-RWKV blocks—structures that combine inverted residual layers with RWKV’s time- and channel-mixing mechanisms. This design allows the model to efficiently extract both local and global information. To complement the encoding process, a cross-channel mix (CCM) module is embedded within the skip connections, reinforcing multiscale feature fusion between encoder and decoder. The architecture demonstrates superior segmentation performance on multiple datasets, including Synapse, ACDC, ISIC 2017, and CVC-ClinicDB, achieving state-of-the-art accuracy while maintaining a lightweight footprint and high inference speed.

Building upon the success of RWKV-UNet, subsequent work introduced Med-URWKV \cite{liu2025medurwkv}, a purely RWKV-based model that eliminates convolutional operations altogether. Med-URWKV adopts a Vision-RWKV encoder pretrained on ImageNet and integrates it into a U-Net style architecture for downstream segmentation tasks. By leveraging pretrained weights, the model benefits from stronger initialization, faster convergence, and better generalization. Compared to models trained from scratch, Med-URWKV consistently yields comparable or superior results across several medical segmentation benchmarks. This pretrained setup significantly reduces the annotation burden and computational cost, making Med-URWKV particularly attractive for low-resource settings or rapid deployment scenarios in clinical environments.

Finally, recent research has explored forward-looking directions that extend RWKV’s architectural capabilities. One notable study proposed QRWKV, a quantum-enhanced RWKV model that incorporates variational quantum circuits into the channel-mixing module to improve expressivity and uncertainty modeling. Applied to small-scale medical classification datasets such as RetinaMNIST and MedMNIST, QRWKV achieved competitive accuracy while exhibiting robustness to noisy or low-contrast inputs. These developments suggest that the RWKV framework is not only efficient and scalable but also extensible into hybrid quantum-classical computing environments. As the field moves forward, RWKV-based models are expected to play a key role in designing compact, high-performance, and interpretable neural networks for both medical and industrial image analysis.

\section{Method}

\subsection{Overall Architecture}
\label{subsec:overall}

MD-RWKV-UNet is a dynamic encoder architecture designed to integrate fine-grained local features with adaptive global context modeling, addressing the challenges of multi-organ segmentation. While following the general U-Net structure with progressive downsampling and hierarchical feature extraction, it introduces a stage-wise encoding framework enhanced by MD-RWKV blocks. Each stage increases semantic abstraction through stacked blocks, beginning with spatial downsampling and followed by within-scale refinement. To maintain coherence across scales, a CrossStageFusion module aggregates features from all preceding stages using dual-attention alignment, enabling the network to adaptively integrate multi-level information.

At the core of the encoder, each MD-RWKV block consists of two branches: a local branch using depthwise separable convolution for efficient texture encoding, and a dynamic branch combining DeformableShift and RWKV-based accumulation. DeformableShift adjusts sampling positions based on content, improving sensitivity to geometric variation, while RWKV accumulates long-range dependencies recursively with exponential decay, capturing global context efficiently. The two branches are fused, regularized via DropPath, and stabilized with residual connections.

To further enhance scale adaptability, we embed SKAttention in early encoder stages, enabling dynamic receptive field adjustment based on spatial content. The final stage normalizes deep features and applies global average pooling to produce a compact image-level descriptor, which is passed to a classification head for prediction. This design forms a lightweight yet expressive encoder capable of robust segmentation across organs with diverse sizes and shapes.

\subsection{MD-RWKV block: Multi-Path Dynamic Residual Block}
\label{subsec:advanced-sde}

The MD-RWKV block is a powerful feature modeling unit that combines multi-path dynamic perception, local convolutional enhancement, and residual connectivity. 
\\
\hspace*{1em}
Given an input feature map $X \in \mathbb{R}^{B \times C \times H \times W}$, where $B$ is the batch size, $C$ the number of channels, and $H \times W$ the spatial resolution, the MD-RWKV block first applies normalization—either LayerNorm or BatchNorm—to reduce inter-sample and inter-channel distribution bias. The normalized feature is then projected into an intermediate dimension $C'$ via a $1 \times 1$ convolution:
\begin{equation}
X' = \mathrm{Conv}_{1 \times 1}(\mathrm{Norm}(X)).
\end{equation}
\\
\hspace*{1em}
The transformed feature $X'$ is processed through two parallel branches: (I) a local convolutional enhancement path and (II) a dynamic spatial attention path. In the local path, the MD-RWKV block employs depthwise separable convolutions to efficiently capture local dependencies. Each channel undergoes a separate spatial convolution, followed by a pointwise convolution that integrates inter-channel information. This decomposition reduces parameters and computation while preserving locality-aware expressiveness.
\\
\hspace*{1em}
In the dynamic path, $X'$ is first processed by the DeformableShift module, which introduces learned spatial offsets for localized spatial adaptivity. The deformed features are then passed through a gating mechanism that computes three components: the key $k$, the value $v$, and the receptance gate $\tau$. These are used as inputs to the RWKV accumulation operation, a core modeling mechanism that performs exponentially weighted aggregation over positions:
\begin{equation}
y(t, c) = \sum_{\tau=0}^{t} e^{w_c (\tau - t)} \cdot (u_c + k(\tau, c)) \cdot v(\tau, c),
\end{equation}
where $w_c$ is a learnable temporal decay coefficient for channel $c$, and $u_c$ is a bias term used to stabilize early activations. This exponential memory mechanism allows the MD-RWKV block to model long-range dependencies with smooth decay, while avoiding the quadratic complexity of traditional attention. The RWKV path thus supports efficient global context modeling with only linear complexity $\mathcal{O}(T)$, where $T$ is the number of positions.
\\
\hspace*{1em}
The output from the RWKV path is normalized and projected back to the intermediate channel dimension $C'$ using a linear layer. During training, DropPath is applied to introduce stochastic depth regularization, enhancing generalization by randomly disabling entire paths. The outputs of the local and dynamic paths are concatenated and fused via a $1 \times 1$ projection. Finally, a residual connection is applied to combine the fused output with the original input:
\begin{equation}
F = \mathrm{Concat}(\mathrm{LocalConv}(X'), \mathrm{RWKV}(X')).
\end{equation}
\begin{equation}
\mathrm{Output} = X + \mathrm{DropPath}\left( \mathrm{Conv}_{1 \times 1}(F) \right).
\end{equation}
\\
\hspace*{1em}
The MD-RWKV block models local correlations through depthwise convolution and captures global dependencies through the RWKV mechanism, which efficiently aggregates context across spatial positions. This design greatly enhances modeling capacity for dynamic local patterns and long-range spatial interactions while maintaining low overhead.

\subsection{CrossStageFusion: Dual Attention Feature Alignment}
\label{subsec:crossstagefusion}

The CrossStageFusion module addresses the semantic and spatial misalignment challenges associated with fusing low-level and high-level features in deep convolutional networks. As the network progresses, deeper layers produce increasingly abstract semantic features with reduced spatial resolution, while shallower layers retain rich local textures and boundary information. 
\\
\hspace*{1em}
Let $F_l \in \mathbb{R}^{B \times C_l \times H \times W}$ and $F_h \in \mathbb{R}^{B \times C_h \times H \times W}$ denote the low- and high-level feature maps, respectively. To ensure spatial alignment, the two are concatenated along the channel dimension:
\begin{equation}
F_{\text{cat}} = \mathrm{Concat}(F_l, F_h).
\end{equation}
This combined representation retains both texture-rich and semantically abstract information, forming a unified basis for attention modeling.
\\
\hspace*{1em}
CrossStageFusion applies both channel attention and spatial attention. The channel attention branch uses two consecutive $1 \times 1$ convolutions with an intermediate ReLU to extract weighting coefficients for each branch:
\begin{equation}
(\alpha_l, \alpha_h) = \mathrm{Split}(\phi(F_{\text{cat}})),
\end{equation}
where $\phi(x)$ denotes two-layer 1×1 convolution with ReLU activation in between. $\alpha_l, \alpha_h \in \mathbb{R}^{B \times 1 \times H \times W}$ are the channel-wise weights for the low and high-level features, respectively.
\\
\hspace*{1em}
The spatial attention branch independently models positional importance. It computes the global average pooling and max pooling of $F_{\text{cat}}$ along the channel dimension, concatenates them, and applies a $7 \times 7$ convolution followed by a sigmoid activation:
\begin{equation}
s = \sigma\left( \mathrm{Conv}{7 \times 7} \left( \mathrm{Pool}{\text{cat}}(F_{\text{cat}}) \right) \right),
\end{equation}
where $s \in \mathbb{R}^{B \times 1 \times H \times W}$ acts as a soft spatial attention mask.
\\
\hspace*{1em}
The final fused output is computed by applying both attention weights to the input features:
\begin{equation}
F_{\text{out}} = (\alpha_l \cdot s) \odot F_l + (\alpha_h \cdot (1 - s)) \odot F_h,
\end{equation}
where $\cdot$ denotes broadcast multiplication and $\odot$ element-wise product. The first term emphasizes low-level details in relevant regions, while the second term enhances semantic abstraction where appropriate.
\\
\hspace*{1em}
CrossStageFusion serves as an adaptive re-weighting mechanism across scales. Channel attention adjusts inter-feature relevance, suppressing redundancy and enhancing discrimination, while spatial attention highlights informative locations based on task relevance. This dual attention strategy ensures consistency and alignment in fused representations.

\begin{table*}[ht]
\centering
\caption{Performance comparison on Synapse and ACDC datasets. Metrics are shown as Dice Similarity Coefficient (↑) / HD95 (↓).}
\label{tab:Tab1}
\begin{tabular}{l|c|c}
\toprule
\textbf{Model} & \textbf{Synapse (DSC / HD95)} & \textbf{ACDC (DSC / HD95)} \\
\midrule
U-Net \cite{Ronneberger2015} & 76.85 / 39.70 & 85.98 / 19.31 \\
Att-UNet \cite{Oktay2018} & 77.77 / 36.02 & 87.55 / 18.74 \\
TransUNet \cite{chen2021} & 77.48 / 31.69 & 89.71 / 17.16 \\
MixedUNet \cite{liu2022} & 78.59 / 26.59 & 89.84 / 15.82 \\
CoTr \cite{xie2021} & 78.56 / 24.05 & 89.97 / 14.59 \\
SwinUNet \cite{cao2022} & 79.13 / 21.55 & 90.00 / 13.23 \\
TransDeepLab \cite{azad2022} & 80.16 / 21.25 & 90.11 / 11.98 \\
SDAUT \cite{huang2022} & 80.67 / 25.59 & 90.23 / 10.67 \\
Hiformer \cite{heidari2023} & 80.69 / 19.14 & 93.65 / 9.41 \\
PVT-GCASCADE \cite{titoriya2023} & 81.06 / 20.23 & 90.45 / 8.26 \\
VM-UNet \cite{ruan2024} & 81.08 / 19.21 & 91.12 / 7.03 \\
nn-UNet \cite{isensee2018} & 82.35 / 18.76 & 91.41 / 6.58 \\
Trans-CASCADE \cite{rahman2024} & 82.68 / 17.34 & 91.63 / 5.84 \\
PVT-EMCAD \cite{rahman2024} & 83.63 / 15.68 & 92.12 / 4.57 \\
xLSTM-VMUNet \cite{fang2024} & 83.74 / 15.59 & 92.14 / 3.12 \\
RWKV-UNet \cite{jiang2025} & 84.02 / 16.09 & 92.17 / 2.75 \\
\textbf{MD-RWKV-UNet} & \textbf{85.07 / 14.67} & \textbf{93.15 / 1.77} \\
\bottomrule
\end{tabular}
\end{table*}

\begin{table*}[ht]
\centering
\caption{Task-wise comparison of RWKV-UNet and MD-RWKV-UNet on Synapse and ACDC datasets. Metrics are shown as Dice Similarity Coefficient (↑) / HD95 (↓).}
\label{tab:Tab2}
\begin{tabular}{ll|c|c}
\toprule
\textbf{Dataset} & \textbf{Task} & \textbf{RWKV-UNet (DSC / HD95)} & \textbf{MD-RWKV-UNet (DSC / HD95)} \\
\midrule
Synapse & Aorta        & 89.53 / 9.39  & 92.05 / 7.86  \\
Synapse & Gallbladder  & 68.94 / 12.22 & 73.55 / 17.62 \\
Synapse & Kidney (L)   & 87.63 / 7.14  & 90.57 / 7.36  \\
Synapse & Kidney (R)   & 84.07 / 19.84 & 86.83 / 8.14  \\
Synapse & Liver        & 95.57 / 15.23 & 98.44 / 33.72 \\
Synapse & Pancreas     & 69.38 / 16.91 & 62.48 / 15.71 \\
Synapse & Spleen       & 90.95 / 27.19 & 94.73 / 9.17  \\
Synapse & Stomach      & 86.09 / 20.79 & 81.89 / 17.80 \\
ACDC    & RV           & 90.86 / 6.14  & 91.68 / 3.16  \\
ACDC    & Myocardium   & 88.72 / 1.06  & 89.50 / 1.13  \\
ACDC    & LV           & 96.92 / 1.04  & 98.28 / 1.04  \\
\bottomrule
\end{tabular}
\end{table*}

\begin{figure*}[htbp]
\centering
\includegraphics[width=\textwidth]{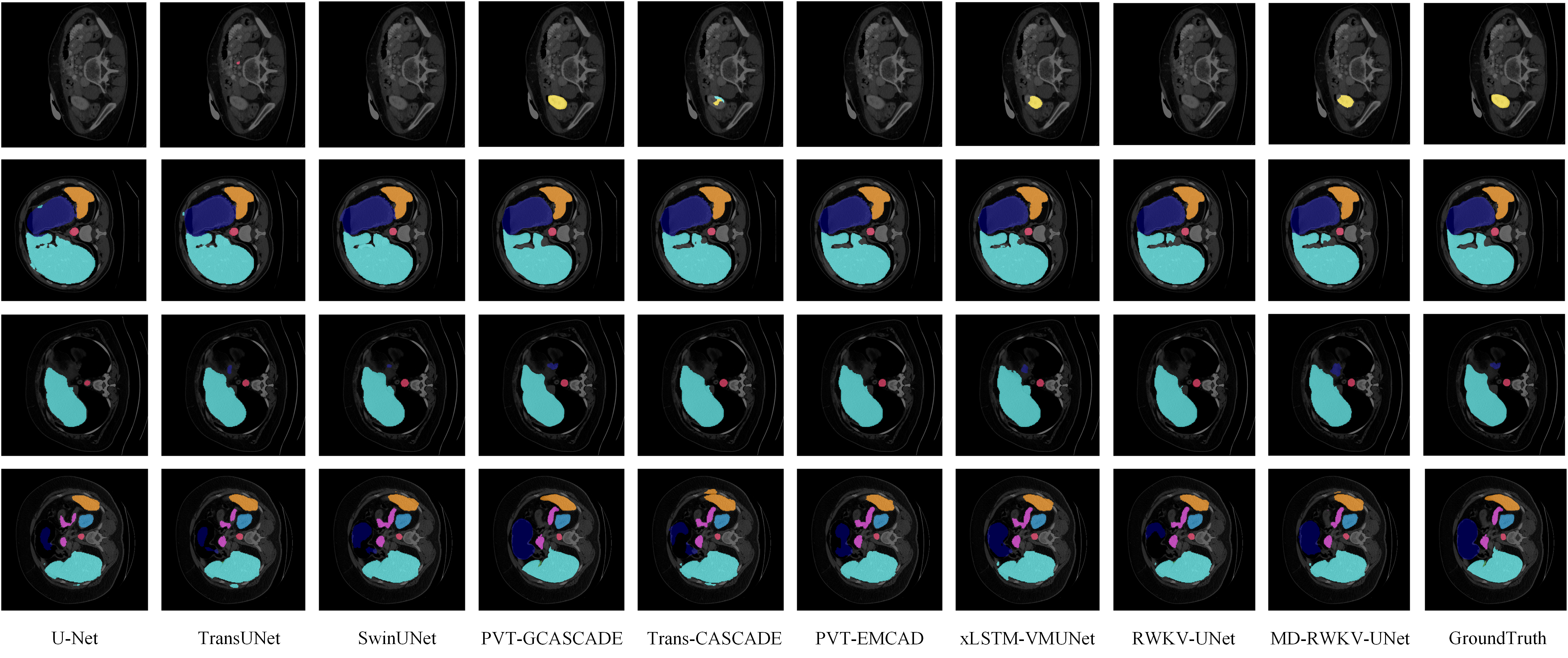}
\caption{A qualitative comparison with previous SOTA methods on the Synapse dataset.}
\label{fig:Fig2}
\end{figure*}

\section{Experiments}

\subsection{Datasets}
\label{datasets}
The Synapse Multi-organ Segmentation dataset, released as part of the 2015 MICCAI Multi-Atlas Labeling Beyond the Cranial Vault Challenge, is a widely used benchmark for abdominal organ segmentation. It contains 30 contrast-enhanced abdominal CT scans, each with manual annotations for 8 organs: aorta, gallbladder, left kidney, right kidney, liver, pancreas, spleen, and stomach. The images are of variable slice thickness and resolution, reflecting realistic clinical scenarios. Due to its challenging anatomical variations and limited data size, Synapse has become a standard testbed for evaluating generalizability and robustness in medical image segmentation models.

The ACDC (Automated Cardiac Diagnosis Challenge) dataset, introduced at MICCAI 2017, is a widely adopted benchmark for cardiac structure segmentation in cine-MRI. It consists of 100 short-axis cardiac MRI scans, each annotated for three anatomical structures: the left ventricle (LV), right ventricle (RV), and myocardium (MYO). The dataset includes scans from five subgroups representing various pathological and normal conditions (e.g., myocardial infarction, dilated cardiomyopathy), ensuring diversity in anatomical variations. Each case contains end-diastolic (ED) and end-systolic (ES) phases, making it suitable for both semantic segmentation and functional analysis. The ACDC dataset is widely used to evaluate the performance, robustness, and generalization ability of cardiac segmentation models.

\subsection{Implementation Details}
\label{details}

All experiments were implemented in PyTorch using our proposed MD-RWKV-UNet model and conducted on a single NVIDIA vGPU with 48GB of memory. Input images are resized to 224×224. Training is performed for 30 and 150 epochs on Synapse and ACDC datasets, respectively. Batch sizes are set to 24. The initial learning rates are 1e-3 (Synapse) and 5e-4 (ACDC), with minimum learning rates of 0. We employ the AdamW optimizer with a CosineAnnealingLR scheduler. Additionally, test-time augmentation (TTA) is applied for Synapse and ACDC evaluations.

\subsection{Results}
\label{results}

We comprehensively evaluate MD-RWKV-UNet and compare it against a broad range of state-of-the-art (SOTA) segmentation models. As shown in Table~\ref{tab:Tab1}, our model achieves the best overall performance across both datasets, with an average Dice score of 85.07 and HD95 of 14.67 on Synapse, and 93.15 and 1.77 on ACDC, outperforming competitive baselines such as xLSTM-VMUNet, PVT-EMCAD, and RWKV-UNet. Notably, compared to its direct predecessor RWKV-UNet, MD-RWKV-UNet improves the average Dice score by 1.05 and reduces HD95 by 1.42 on Synapse, while also achieving a substantial HD95 reduction of 0.98 on ACDC. These results demonstrate the strength of our design in achieving both accurate segmentation and precise boundary delineation, particularly in anatomically complex and multi-scale scenarios.

\begin{figure*}[htbp]
\centering
\includegraphics[width=\textwidth]{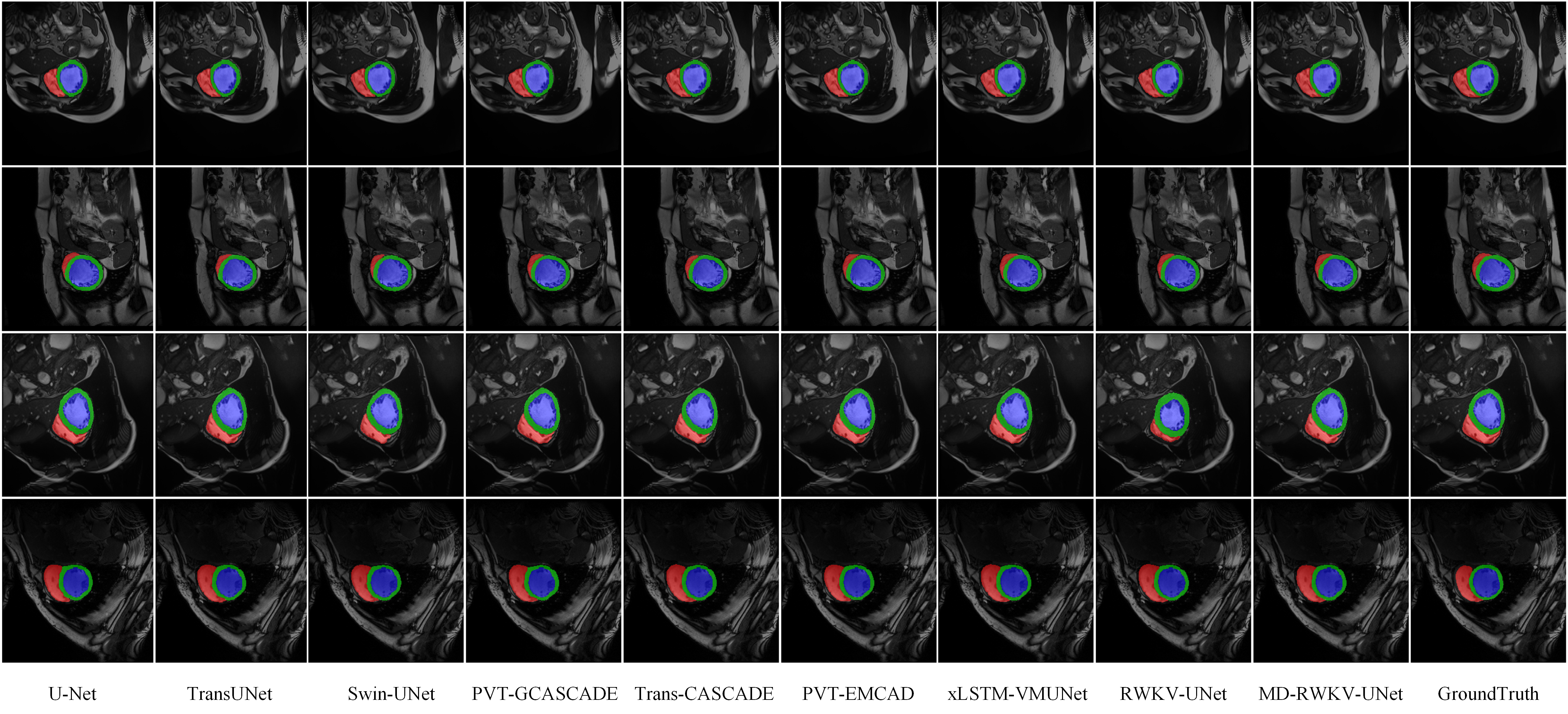}
\caption{A qualitative comparison with previous SOTA methods on the ACDC dataset.}
\label{fig:Fig3}
\end{figure*}

To further assess model behavior in individual organ-level tasks, we conduct a task-wise comparison between MD-RWKV-UNet and RWKV-UNet, as summarized in Table~\ref{tab:Tab2}. The proposed model consistently improves segmentation accuracy across most organs. On Synapse, significant gains are observed in large and structurally diverse organs such as the aorta (from 89.53 to 92.05 Dice), left kidney (from 87.63 to 90.57), and spleen (from 90.95 to 94.73). MD-RWKV-UNet also demonstrates remarkable improvements in boundary precision, as reflected in HD95 reductions—for instance, from 27.19 to 9.17 in spleen and from 19.84 to 8.14 in the right kidney. Notably, even small structures like the gallbladder benefit from the proposed model (+4.6 Dice), although pancreas and stomach exhibit slight performance trade-offs, which may be attributed to their low contrast and high shape variability. On ACDC, consistent gains are achieved across all cardiac structures, including the left ventricle (Dice +1.36), right ventricle (Dice +0.82), and myocardium (Dice +0.78), with significant improvements in boundary accuracy for the right ventricle (HD95 reduced from 6.14 to 3.16). Figure~\ref{fig:Fig2} and Figure~\ref{fig:Fig3} provide a visual illustration of the segmentation results.

\subsection{Ablation Study}
\label{results}

\begin{table*}[ht]
\begin{center}
\begin{tabular}{|l|c|c|c|c|c|}
\hline
\textbf{Ver.} & \textbf{SKAttention} & \textbf{DeformableShift} & \textbf{CrossStageFusion} & \textbf{Synapse (DSC / HD95)} & \textbf{ACDC (DSC / HD95)} \\
\hline\hline
Ver 1 & \ding{55} & \ding{55} & \ding{55} & 84.02 / 16.09 & 92.17 / 2.75 \\
Ver 2 & \ding{51} & \ding{55} & \ding{55} & 84.19 / 15.99 & 92.28 / 2.66 \\
Ver 3 & \ding{55} & \ding{51} & \ding{55} & 84.37 / 15.73 & 92.45 / 2.48 \\
Ver 4 & \ding{55} & \ding{55} & \ding{51} & 84.56 / 15.48 & 92.63 / 2.33 \\
Ver 5 & \ding{51} & \ding{51} & \ding{55} & 84.74 / 15.22 & 92.81 / 2.18 \\
Ver 6 & \ding{51} & \ding{55} & \ding{51} & 84.89 / 14.96 & 92.97 / 2.04 \\
Ver 7 & \ding{55} & \ding{51} & \ding{51} & 85.03 / 14.78 & 93.09 / 1.89 \\
Ver 8 & \ding{51} & \ding{51} & \ding{51} & \textbf{85.07 / 14.67} & \textbf{93.15 / 1.77} \\
\hline
\end{tabular}
\end{center}
\caption{Ablation study of key modules (SKAttention, DeformableShift, CrossStageFusion) in MD-RWKV-UNet across Synapse and ACDC datasets.}
\label{tab:Tab3}
\end{table*}

To further understand the contributions of each architectural component in MD-RWKV-UNet, we conduct an ablation study on the Synapse and ACDC datasets, evaluating the impact of three key modules: SKAttention, DeformableShift, and CrossStageFusion. Table~\ref{tab:Tab3} reports the results of eight model variants, incrementally enabling each module to assess their individual and combined contributions.

Enabling SKAttention alone (Ver 2) yields consistent, albeit modest improvements over the baseline (Ver 1), reflecting its ability to adapt receptive fields to organ-specific scale variations. This is particularly beneficial for segmenting both large organs (e.g., liver, spleen) and smaller structures (e.g., pancreas, gallbladder) in abdominal CT scans.

Introducing DeformableShift (Ver 3) produces greater gains, indicating its strength in capturing local spatial deformations and anatomical variability. When combined with SKAttention (Ver 5), performance further improves, confirming the complementarity of scale-aware and spatially adaptive modeling.

CrossStageFusion (Ver 4) also improves results by effectively aligning multi-scale features across encoding depths. Its synergy with either SKAttention (Ver 6) or DeformableShift (Ver 7) enhances boundary precision and semantic consistency. The full model (Ver 8) achieves the highest performance across both datasets, validating the integrated design.

In summary, each module addresses a distinct challenge in multi-organ segmentation: SKAttention tackles scale heterogeneity, DeformableShift improves spatial flexibility, and CrossStageFusion enhances hierarchical feature coherence. Their integration yields a robust and generalizable architecture for anatomically diverse segmentation tasks.

\section{Conclusion}

In this work, we propose MD-RWKV-UNet, a dynamic encoder framework designed for multi-organ medical image segmentation. To address the challenges of anatomical variability and scale diversity, the model integrates three key components: the MD-RWKV block for deformable spatial modeling and efficient global context aggregation, Selective Kernel Attention for adaptive receptive field selection, and a cross-stage dual-attention fusion mechanism for multi-level feature alignment. Extensive experiments on Synapse and ACDC demonstrate that MD-RWKV-UNet consistently outperforms state-of-the-art baselines, particularly in boundary precision and small-organ segmentation. Moving forward, we believe MD-RWKV-UNet provides a strong foundation for future exploration in dynamic medical image analysis.

\bibliography{aaai2026}

\end{document}